\pdfoutput=1
\documentclass[11pt]{article}

\usepackage[]{acl}

\usepackage{times}
\usepackage{latexsym}

\usepackage[T1]{fontenc}
\usepackage[utf8]{inputenc}

\usepackage{microtype}

\usepackage{graphicx}
\usepackage{multirow}
\usepackage{adjustbox}
\usepackage{booktabs}

\usepackage{caption}
\usepackage{subcaption}

\definecolor{airforceblue}{rgb}{0.36, 0.54, 0.66}
\definecolor{blue(ncs)}{rgb}{0.0, 0.53, 0.74}
\definecolor{blush}{rgb}{0.87, 0.36, 0.51}
\definecolor{softgreen}{rgb}{0.7, 0.9, 0.7}
\definecolor{darkgreen}{rgb}{0.0, 0.5, 0.0} 
\definecolor{darkred}{rgb}{0.5, 0.0, 0.0} 

\title{CRAFT: Extracting and Tuning Cultural Instructions from the Wild}

\author{Bin Wang\textsuperscript{$\heartsuit$}, 
Geyu Lin\textsuperscript{$\heartsuit$},
Zhengyuan Liu\textsuperscript{$\heartsuit$}, 
Chengwei Wei\textsuperscript{$\heartsuit$},
Nancy F. Chen\textsuperscript{$\heartsuit,\dag$}
\\
\textsuperscript{$\heartsuit$}Institute for Infocomm Research (I$^2$R), A*STAR, Singapore\\
\textsuperscript{$\dag$}Centre for Frontier AI Research (CFAR), A*STAR, Singapore\\
\texttt{wang\_bin@i2r.a-star.edu.sg} \\
}

\begin{document}
\maketitle

\begin{abstract}

    Large language models (LLMs) have rapidly evolved as the foundation of various natural language processing (NLP) applications. Despite their wide use cases, their understanding of culturally-related concepts and reasoning remains limited. Meantime, there is a significant need to enhance these models' cultural reasoning capabilities, especially concerning underrepresented regions. This paper introduces a novel pipeline for extracting high-quality, culturally-related instruction tuning datasets from vast unstructured corpora. We utilize a self-instruction generation pipeline to identify cultural concepts and trigger instruction. By integrating with a general-purpose instruction tuning dataset, our model demonstrates enhanced capabilities in recognizing and understanding regional cultural nuances, thereby enhancing its reasoning capabilities. We conduct experiments across three regions: Singapore, the Philippines, and the United States, achieving performance improvement of up to 6\%. Our research opens new avenues for extracting cultural instruction tuning sets directly from unstructured data, setting a precedent for future innovations in the field.\footnote{Our models and datasets are available for future research at~\url{https://github.com/SeaEval/CRAFT}.}    

\end{abstract}

\section{Introduction}

    Large language models (LLMs) like ChatGPT~\cite{achiam2023gpt}, Claude, and Gemini~\cite{reid2024gemini} have demonstrated their proficiency in managing diverse tasks related to semantic understanding and text generation. Beyond acting as general task-solvers, the ability of LLMs to understand and reason with cultural nuances could play a crucial role in generating precise and personalized responses to benefit broader communities~\cite{tao2023auditing,wang2023seaeval,adilazuarda2024towards}.

    Culture is a comprehensive concept encompassing traditions, customs, beliefs, values, and social norms, all deeply rooted in historical contexts and continuously evolving over time. It is also intrinsically linked to languages and dialects, which can be sparsely represented in available resources.

    In the domain of LLMs, which initially train on vast amounts of unlabeled data, knowledge is systematically captured and structured through data-driven techniques. With limited model sizes, knowledge that occurs infrequently is often less effectively captured compared to more frequently occurring information~\cite{kaplan2020scaling}. Additionally, the predominance of English in the pre-training corpus inherently biases these models towards Western perspectives, a consequence of the over-representation of English-language sources. This bias means that cultural concepts may be inadequately captured, especially for under-represented regions~\cite{masoud2023cultural}. Consequently, LLMs struggle to effectively adapt to and represent diverse cultural concepts due to these inherent limitations in their training data.

    \begin{figure*}[t]
         \centering
             \includegraphics[width=1.00\textwidth]{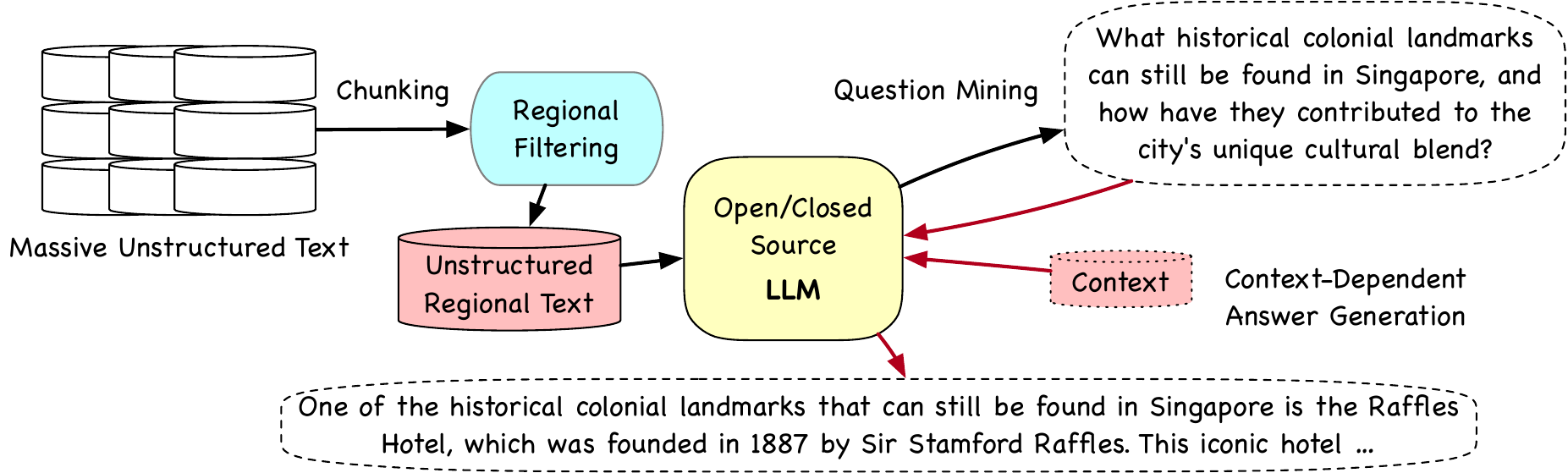}
            \caption{
            The CRAFT method involves creating instruction datasets tailored for culturally rich instruction by processing extensive unstructured data with large language models (LLMs). These specialized cultural instructions are then employed to improve the ability of LLMs to reason within cultural contexts through instruction fine-tuning.
            }
            \label{fig:framework}
    \end{figure*}

    Expanding the cultural reasoning capabilities of LLMs could potentially be achieved by pre-training them on corpora from diverse languages. However, this approach is still expansive and challenging due to the difficulty in obtaining high-quality multilingual datasets~\cite{bai2023qwen,sea_lion_2023}. Meanwhile, instruction fine-tuning could more directly impact end-user applications. However, the development of cultural instruction tuning sets is limited due to the high costs associated with collecting culturally relevant instruction sets, along with challenges in ensuring their quality and diversity.

    In this study, we pioneer the study of deriving instruction tuning sets from unlabeled corpora. Initially, we use keyword filtering to isolate culturally relevant concepts from a vast corpus containing over 600 billion English tokens. Subsequently, we then utilized these selected regional text segments to prompt LLMs for both questions and answers. Our evaluations focus on the context of Singapore and extend to the Philippines and the US. Our experiments utilize the \emph{Mistral-7B} model, combining general instructions with our specifically curated cultural instruction set. We observe performance improvement of up to 6\% while maintaining intelligence in general subject knowledge as assessed by the MMLU dataset. Additionally, we also analyzed the impact of answer sources and the ratio of cultural instructions. Both the model and the curated cultural instruction-tuning dataset are made available for future research.

\section{Related Work}

    Our work aims to improve the cultural reasoning capability of LLMs. The capability of LLMs can be refined in their training schemes, such as pre-training, instruction tuning, and RLHF preference optimization. Current efforts often concentrate on fine-tuning using specific datasets derived from diverse multilingual corpus sources~\cite{lin2023taiwan,abbasi2023persianllama,pipatanakul2023typhoon}. Yet, this approach is costly in training and lacks transparency regarding the extent of cultural concepts incorporated into the model. \citet{li2024culturellm} propose to leverage a set of opinion questions to gather views towards different cultural groups, which is then used to finetune a model. However, despite the efforts at data augmentation, the scope of these questions remains limited and fails to encompass a wide range of cultural dimensions. Therefore, in our research, we focus on extracting a wide variety of cultural instructions from large unlabeled corpora, ensuring guaranteed diversity.

    Given the complexity of sourcing high-quality instructional data, LLMs are employed to create synthetic question-answering pairs~\cite{wang2022self,wang-etal-2023-instructive} and dialogue data through iterative processes~\cite{ding-etal-2023-enhancing}. However, these approaches tend to concentrate on generating general instructions and dialogues, lacking the capability to produce culturally rich instructions. Prompting LLMs to directly generate cultural concepts is also challenging, as these concepts are sparsely distributed across various resources.
    
    \begin{table*}[t]
        \centering
        \begin{adjustbox}{width=1.00\textwidth,center}
        \begin{tabular}{ l | c   c   c | c }
        \toprule
         \textbf{Models} &\textbf{SG-Eval} & \textbf{PH-Eval} & \textbf{US-Eval} & \textbf{MMLU} \\  \hline \hline
         \multicolumn{1}{l}{{\textbf{General LLMs}}} \\\hline\hline
         
         \emph{ChatGPT-3.5} & 64.4 & 58.4 & 74.9 & 67.5 \\
         \emph{LLaMA-3-8B-Instruct}~\cite{MetaLLama3} & 62.1 & 54.6 & 69.8 & 62.5  \\
         \emph{LLaMA-2-7B-Chat}~\cite{touvron2023llama} & 39.8 & 35.4 & 50.1 & 44.5 \\
         \emph{Mistral-7B-Instruct-v0.2} & 62.1 & 48.6 & 60.9 & 58.1 \\
         \hline \hline
    
        \multicolumn{5}{l}{\textbf{Base Model}: \emph{Mistral-7B-Instruct-v0.2}~\cite{jiang2023mistral}} \\\hline\hline
         Tuning w/ OpenHermes-2.5 & 62.6 & 46.4 & 65.5 & 58.5 \\\hline
         CRAFT$_{sg}$  & \textbf{68.3} / 64.2 & 46.2 / 44.8 & 65.2 / 64.6 & 59.8 / 58.4 \\
         CRAFT$_{ph}$  & 64.3 / 63.7 & \textbf{49.2} / 44.6 & 65.4 / 64.7 & 60.0 / 59.4 \\
         CRAFT$_{us}$  & 65.3 / 63.2 & 48.4 / 44.8 & \textbf{67.1} / 63.5 & 60.3 / 59.7 \\
        \bottomrule
        \end{tabular}
        \end{adjustbox}
        \caption{The main results for general LLMs and our model across three cultural evaluation datasets and the MMLU dataset, which assesses general knowledge. For our results, "-/-" denotes the scores for context-dependent answers and context-free answers, respectively.} 
        \label{tab:results}
    \end{table*}

\section{Methodology}

    We introduce the \textbf{CRAFT} (\textbf{C}ultural \textbf{R}e\textbf{A}soning with Instruction \textbf{F}ine-\textbf{T}uning) method, designed to synthesize cultural instructions from a massive, unlabeled English corpus. The methodology is detailed in the following steps, as illustrated in Figure~\ref{fig:framework}.

    \noindent\textbf{Selective Data Extraction}. We utilize SlimPajama~\cite{cerebras2023slimpajama} as our primary data source, which comprises an English corpus containing over 600 billion tokens. Given the sparse distribution of culturally relevant concepts within this vast dataset, and to manage the processing burden effectively, we propose an efficient filtering process using keywords to identify and extract culture-related concepts.

    Specifically, we curate a collection comprising a minimum of 150 words to represent each region. We then segment the documents into chunks no larger than 512 tokens each. From these segments, we retain only those text chunks that include at least two regional keywords, such as "National Day Parade" and "Merlion" for Singapore. Through analyzing a subset of over 200 billion English tokens, we successfully extracted 35,000 text segments for Singapore, 25,000 for the Philippines, and 35,000 for the US.
    
    \noindent\textbf{Automated Question Creation}. Given the text chunks rich in cultural and local content, we prompt an off-the-shelf LLM to generate questions specifically related to each chunk, focusing on the cultural and regional concepts mentioned.

    \noindent\textbf{Answer Production}. To collect responses for the generated questions, we employ two approaches: 1) context-dependent answer generation, where the given context is provided to LLMs when forming answers, as shown in Figure~\ref{fig:framework}, and 2) context-free answer generation, allowing for responses that are more creative and less tailored to the immediate context. For automatic question creation and context-dependent answer generation, we utilize \emph{Zephyr-7B-Beta}~\cite{tunstall2023zephyr}. For context-free answer generation, we employ \emph{ChatGPT-3.5}.

    \noindent\textbf{Hybrid Instruction Tuning}. After developing the cultural instructions, we compiled at least 20,000 instructions for each specified region. To ensure a balanced capability, we incorporated 50,000 single-round instructions from the OpenHermes-2.5~\cite{OpenHermes_2_5} dataset alongside random sampled 20,000 cultural instructions to fine-tune the \emph{Mistral-7B-Instruct-v0.2}~\cite{jiang2023mistral} model. We adapt LoRA~\cite{hu2021lora} training with a rank of 16 to effectively integrate our instruction tuning knowledge into the model.

\section{Experiments}

    \noindent\textbf{Settings}. For evaluation, we selected four datasets, including three culturally-focused ones: SG-Eval, Ph-Eval, and US-Eval~\cite{wang2023seaeval}. Each of these datasets is a human-crafted collection of multiple-choice questions and answers that probe cultural or regional knowledge. To assess knowledge capabilities in general subjects, we utilize a subset of the MMLU dataset~\cite{hendrycks2020measuring,wang2023seaeval}. To minimize the influence of randomness, we report the average results from five runs, using paraphrased prompt templates.

    \noindent\textbf{Results and Discussion}. Table~\ref{tab:results} shows the main result where we listed four competitive LLMs and our baseline model tuned with 50,000 OpenHermes instructions. First, we witness that \emph{ChatGPT-3.5} and recently released \emph{LLaMA-3-8B-Instruct} are strong in capturing cultural concepts from different regions. Compared to \emph{Mistral-7B} models, LLaMA-3's larger training data and multilingual tokens could help improve the model's capability.

    \begin{table*}[t]
        \centering
        \begin{adjustbox}{width=0.975\textwidth,center}
        \begin{tabular}{ c| l }
        \toprule
        \textbf{Region} & \multicolumn{1}{c}{\textbf{Instruction and Answer}}  \\ \midrule
       
        \multirow{5}{*}{\textbf{SG}} & \textbf{Q}: \emph{What historical significance does the Mandarin Orchard Singapore hold, and how has it been} \\
           & \emph{preserved while undergoing a major refurbishment?} \textbf{A1}: \emph{The Mandarin Orchard Singapore holds} \\ 
           &  \emph{historical significance as it is a landmark building on Orchard Road, with its two distinctive towers} \\
           & \emph{built in 1971 and 1973. ...} \textbf{A2}: \emph{The Mandarin Orchard Singapore, previously known as the Mandarin} \\
           &  \emph{Singapore, holds significant historical and cultural importance as one of Singapore's ...} \\ 
           \midrule

        \multirow{3}{*}{\textbf{SG}} & \textbf{Q}: \emph{How has Kuala Lumpur's mix of Malay, Indian, and Chinese cultures contributed to its unique} \\
           & \emph{experience and creative options for event planners, as mentioned in the context about Kuala Lumpur} \\
           & \emph{being a city of vibrant contrasts?} \textbf{A2}: \emph{Kuala Lumpur's blend of Malay, ...} \\

           \hline

        \multirow{2}{*}{\textbf{PH}} & \textbf{Q}: \emph{How is the Araquio festival celebrated in Nueva Ecija, and what is its traditional month of } \\ 
        & \emph{celebration?} \textbf{A1}: \emph{The Araquio festival is traditionally celebrated in the month of May in Nueva ...} \\\hline

        \multirow{3}{*}{\textbf{US}} & \textbf{Q}: \emph{How did the discovery of gold in California in 1848 impact the transition from Mexican ranching}\\ 

        & \emph{to American farms and towns in the East Bay area, and what were the consequences for the Estudillo} \\

        & \emph{and Peralta families who owned land in the region?} \textbf{A2}: \emph{The discovery of gold in California ...} \\

        \bottomrule
        \end{tabular}
        \end{adjustbox}
        \caption{Synthesized cultural instructions and answers. \textbf{A1} and \textbf{A2} refer to context-dependent and context-free answers, respectively.} 
        \label{tab:examples}
    \end{table*}

    Compared to our baseline, our CRAFT model demonstrates a clear performance improvement across all three regions. Notably, we observe more significant enhancements for Singapore-related questions, improving from 62.6\% to 68.3\%. We attribute this to two main reasons: 1) It is a culturally rich region with mixed and unique cultural origins, and 2) Singaporean concepts are well-documented in English corpora, even though they constitute a small portion. In contrast, cultural knowledge of the US appears more frequently during the pretraining stage, leading to better learning outcomes. Conversely, cultural knowledge from the Philippines is less documented and more dispersed. We encountered similar challenges when collecting culturally related text chunks for it. We found that identifying US text chunks required the least data, followed by Singapore, with the Philippines proving to be the most challenging due to its scattered documentation.

    \begin{figure}[t]
         \centering
             \includegraphics[width=0.42\textwidth]{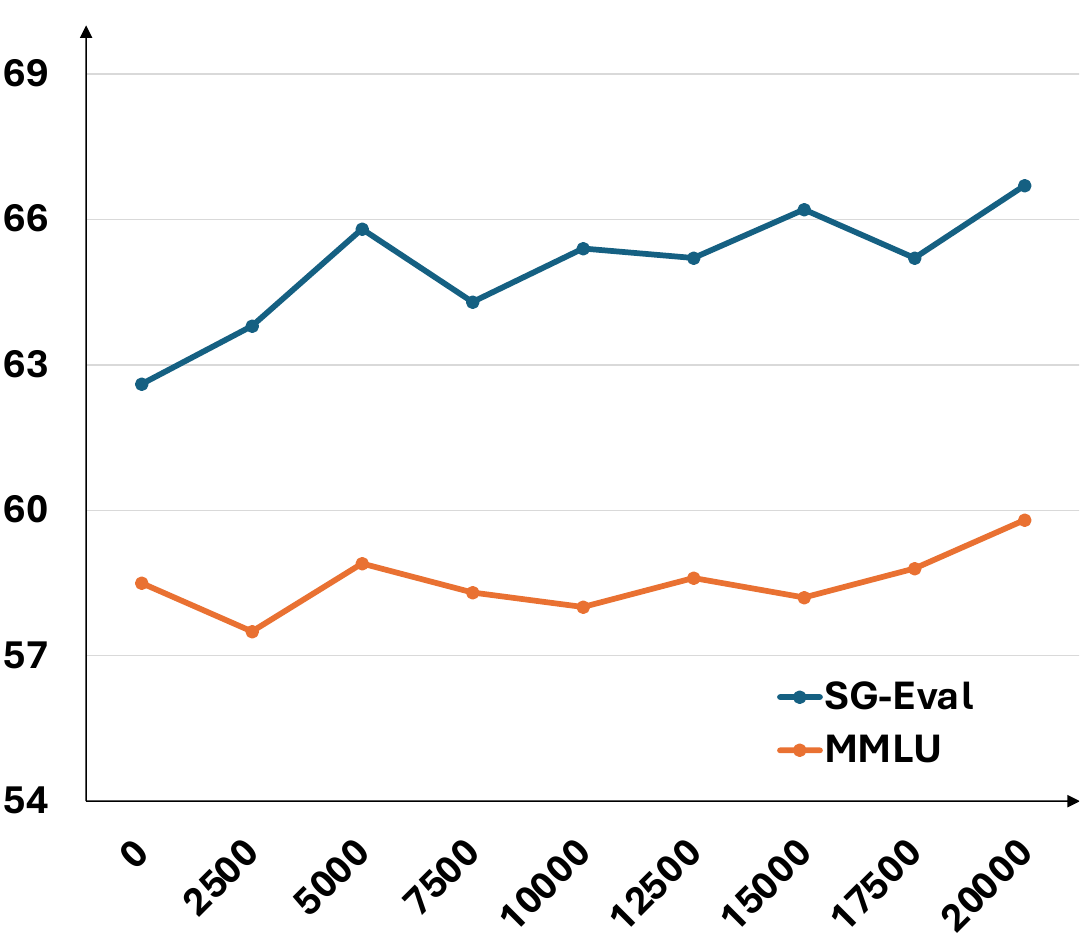}
            \caption{
            Performance on SG-Eval and MMLU dataset. The CRAFT method with different ratios of Singapore cultural instruction data.
            }
            \label{fig:ratio}
    \end{figure}

    Lastly, when comparing context-dependent answers with context-free answers, we found that responses derived from context are consistently more reliable and informative. Consequently, instructions that utilize context-dependent answers consistently yield higher performance gains. While context-free answers could potentially become more informative using advanced models like GPT-4, this enhancement would also benefit context-dependent answers.

    \noindent\textbf{Cultural Instruction Ratios}. In Figure~\ref{fig:ratio}, we investigate the effects of adding varying amounts of cultural instructions to a base of 50,000 general instructions. Cultural instructions are incrementally introduced into the training data at intervals of 2,500 samples. The results indicate that performance on the SG-Eval improves as more culturally related samples are added, suggesting that an increased number of cultural concepts are activated from the pre-training phase and further improved from the instructions. Concurrently, our analysis shows that the general performance as measured by the MMLU datasets remains consistent.

    \noindent\textbf{Instruction Samples}. In this section, we present a series of synthesized cultural instructions and their corresponding responses generated by various methods, as illustrated in Table~\ref{tab:examples}. The samples demonstrate that culturally related questions can be effectively derived from culturally specific content. Context-dependent responses tend to incorporate more factual knowledge from the provided context compared to context-free answers. However, this can also lead to biases based on the limited facts presented. We observe that context-free responses often lack depth in knowledge-intensive instructions due to the model’s inherent limitations.

\section{Conclusion}

    In this paper, we introduce the CRAFT method, designed to synthesize cultural instructions from a vast, unlabeled corpus. We conduct experiments across three regions, with the potential for expansion to additional regions. Our pioneering self-instruction techniques facilitate effective mining from unstructured data sources, enhancing both the diversity and quality of the synthesized instructions compared to previous studies.

\section*{Limitations}

    In this study, we concentrate on mining cultural instructions from English-language corpora. However, it is important to recognize that cultural concepts are often deeply integrated with their respective languages, including those that are primarily spoken. Therefore, to effectively synthesize cultural concepts and instructions, adopting multilingual approaches~\cite{liu2023multilingual,lin2024crossin} is essential to accommodate a broader range of cultural contexts.

\section*{Acknowledgement}

    This work is supported by the National Research Foundation, Singapore under its AI Singapore Programme (AISG Award No: AISG2-GC-2022-005). This research is also supported by the National Research Foundation, Singapore and Infocomm Media Development Authority, Singapore under its National Large Language Models Funding Initiative. Any opinions, findings and conclusions or recommendations expressed in this material are those of the author(s) and do not reflect the views of National Research Foundation, Singapore and Infocomm Media Development Authority, Singapore.

\bibliography{anthology,custom}

\appendix

\section{Keywords and Templates}
\label{sec_app:prompt}

    This section details the keywords and instructions employed for selective data extraction, automated question generation, and answer formulation.

    Regarding cultural keywords, examples are provided below. Our work focuses on the pioneering effort to synthesize cultural instructions. Thus, these keywords can be further refined to enhance performance.

    For keywords, a complete list can be found in our source code.

    \begin{itemize}
        \item \textbf{Singapore}: "HDB flats", "Bukit Merah", "Jurong West", "Orchard Road", "Marina Bay Sands", "Merlion", "Sentosa Island", "CPF (Central Provident Fund)", "Temasek", "Lee Kuan Yew", "BTO (Build-To-Order)", "SCDF (Singapore Civil Defence Force)", "Majulah Singapura (National Anthem)", etc.

        \item \textbf{the Philippines}: "Jeepney", "Barong Tagalog", "Sinulog Festival", "Adobo", "Balut", "OFW (Overseas Filipino Worker)", "Bayanihan", "Boracay", "Ifugao Rice Terraces", "Bahay Kubo", "Tinikling", etc.

        \item \textbf{USA}: "The Statue of Liberty", "Hollywood", "Route 66", "The Grand Canyon", "Mount Rushmore", "The Civil Rights Movement", "Mardi Gras", "The White House", "NASCAR", "Manhattan Project", "Broadway", "Apollo Moon Landing", "Civil War", "Harlem Renaissance", etc.
        
    \end{itemize}
    
    We employ a straightforward prompt for question generation: "You are a chatbot who always generates just one question about Singapore from the given context. Do not generate the answer.". For generating both context-dependent and context-free answers, we use the direct prompt:  "Please answer the following question.". The difference lies in whether the context is provided and the model will decide on how much the answer should rely on the provided context.

\end{document}